\documentclass{article}

\usepackage{PRIMEarxiv}

\usepackage[utf8]{inputenc} 
\usepackage[T1]{fontenc}    
\usepackage{hyperref}       
\usepackage{url}            
\usepackage{booktabs}       
\usepackage{amsfonts}       
\usepackage{nicefrac}       
\usepackage{microtype}      
\usepackage{lipsum}
\usepackage{fancyhdr}       
\usepackage{graphicx}       
\graphicspath{{media/}}     

\usepackage{algorithm}
\usepackage{algorithmic}
\usepackage{xspace}
\usepackage{amsmath}
\usepackage{booktabs}
\usepackage{threeparttable}
\usepackage{makecell}

\hypersetup{
colorlinks=true,
linkcolor=black,
citecolor=black
}

\newcommand{\baby}{S$^2$M$^2$-SAR\xspace}

\title{Semi-supervised Multiscale Matching for SAR-Optical Image}

\author{
  Jingze Gai \\
  Nanyang Technological University \\
  Singapore \\
  \texttt{c240150@e.ntu.edu.sg} \\
   \And
  Changchun Li \\
  Jilin University \\
  Jilin, China \\
  \texttt{changchunli93@gmail.com} \\
}

\begin{document}
\maketitle

\begin{abstract}
  Driven by the complementary nature of optical and synthetic aperture radar (SAR) images, SAR-optical image matching has garnered significant interest. 
Most existing SAR-optical image matching methods aim to capture effective matching features by employing the supervision of pixel-level matched correspondences within SAR-optical image pairs, which, however, suffers from time-consuming and complex manual annotation, making it difficult to collect sufficient labeled SAR-optical image pairs.
To handle this, we design a semi-supervised SAR-optical image matching pipeline that leverages both scarce labeled and abundant unlabeled image pairs and propose a semi-supervised multiscale matching for SAR-optical image matching (\baby). 
Specifically, we pseudo-label those unlabeled SAR-optical image pairs with pseudo ground-truth similarity heatmaps by combining both deep and shallow level matching results, and train the matching model by employing labeled and pseudo-labeled similarity heatmaps.
In addition, we introduce a cross-modal feature enhancement module trained using a cross-modality mutual independence loss, which requires no ground-truth labels. This unsupervised objective promotes the separation of modality-shared and modality-specific features by encouraging statistical independence between them, enabling effective feature disentanglement across optical and SAR modalities.

To evaluate the effectiveness of \baby, we compare it with existing competitors on benchmark datasets. Experimental results demonstrate that \baby not only surpasses existing semi-supervised methods but also achieves performance competitive with fully supervised SOTA methods, demonstrating its efficiency and practical potential.
\end{abstract}

\section{Introduction}
Synthetic Aperture Radar (SAR) images are widely utilized in remote sensing applications because they can capture structural details of the Earth’s surface under all weather and lighting conditions~\cite{rs15030850}. In tasks such as object detection and multimodal image fusion, it is often necessary to match SAR images with corresponding optical remote sensing images. However, achieving accurate alignment between SAR and optical images remains highly challenging because of the significant heterogeneity in their imaging mechanisms. To address this issue, numerous SAR-optical image matching methods have been developed, encompassing both traditional algorithms and deep learning-based approaches~\cite{zhang2025multiresolutionsaropticalremote}.

Traditional SAR-optical matching methods include template-based approaches, which match image patches using mutual information (MI) or normalized cross-correlation (NCC)~\cite{1344163,Ayubi:24,ye2021improvingcoregistrationsentinel1sar}, and feature-based approaches that use hand-crafted descriptors like SIFT~\cite{Lowe2004DistinctiveIF} and its variants (e.g., SAR-SIFT~\cite{6824220}, OS-SIFT~\cite{8272317}). Although effective within single modalities, these traditional methods often fail with SAR-optical pairs due to significant modality differences, radiometric discrepancies, and geometric distortions. 

Recently, deep learning methods have become popular, aiming to learn modality-invariant features by mapping SAR-optical images into a unified feature space through approaches like contrastive learning and diffusion~\cite{10.1007/s11263-020-01359-2,Xu_2024,ye20243mosmultisourcesmultiresolutionsmultiscenes,8898635}.

Despite recent advances in SAR-optical image matching that have improved accuracy and robustness, most state-of-the-art models remain highly data-hungry, relying on large volumes of precisely aligned SAR-optical image pairs. While datasets like SEN1-2~\cite{schmitt2018sen12datasetdeeplearning} have facilitated data-driven learning, the diversity of SAR and optical sensors often necessitates fine-tuning or retraining when applied to new datasets. These models still depend heavily on extensive ground-truth correspondences, which are costly to obtain due to the need for manual annotation (e.g., human-labeled tie-points) or precise sensor co-calibration. As Hughes et al.~\cite{hughes2019semi} observe, the scarcity of large-scale SAR-optical patch datasets is a major barrier to building generalizable matching models.

To address this challenge, semi-supervised learning has emerged as a promising solution to the challenge of limited labeled data in SAR-optical image matching. These approaches leverage large volumes of readily available, unmatched SAR-optical pairs alongside a small set of accurately matched examples. For instance, Hughes et al.~\cite{hughes2019semi} investigated SAR-optical matching under limited supervision, initiating work in this direction. Du et al.~\cite{Du2022ASI} proposed a modality translation framework that simplifies the problem by converting it into a unimodal task. However, most existing semi-supervised methods~\cite{Khurshid_2020} treat unlabeled data merely as auxiliary input for translation, without fully leveraging the coherence of feature distributions across labeled and unlabeled pairs—ultimately limiting matching accuracy.

To better understand the impact of label scarcity, we analyze multilevel feature matching using a U-Net-based template matching model with normalized cross-correlation (NCC). We extract both shallow-level (high-resolution) and deep-level (low-resolution) feature maps. Our findings show that shallow-level matching, while capable of pixel-level precision, is highly sensitive to annotation quality and requires substantial labeled data. In contrast, deep-level features capture broader geometric and semantic context, enabling coarse yet robust matching with significantly fewer labels and faster convergence due to the reduced number of candidate matching positions.

\begin{figure}[h]
  \centering
  \includegraphics[width=\linewidth]{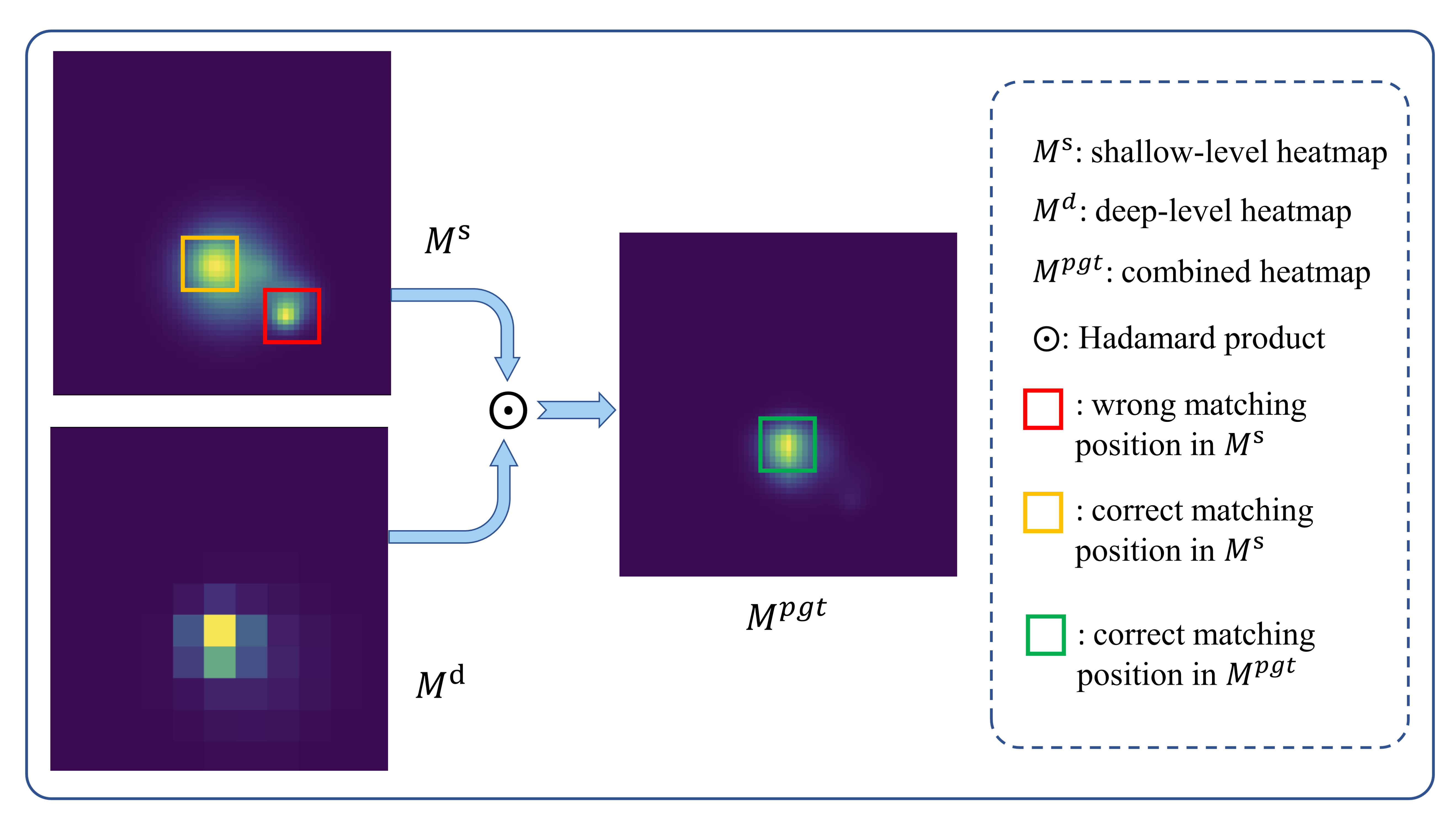}
  \caption{Combining deep-level (robust but low-resolution) and shallow-level (high-resolution but prone to mismatches) similarity heatmaps.}
  \label{fig:multiscale}
\end{figure}

Moreover, we observe that in many instances of mismatching at the shallow level, the similarity score at the correct matching position remains relatively high, albeit slightly lower than incorrect peaks.This indicates that the matching is not entirely erroneous but is dominated by one or a few incorrect predictions with marginally higher scores. These observations underpin our semi-supervised framework, which uses multiscale feature consistency to generate pseudo-labels—leveraging the robustness of deep-level predictions to refine shallow-level matching without heavy annotation costs. Figure~\ref{fig:multiscale} illustrates how combining heatmaps from multiple scales leads to more robust matching results, serving as reliable pseudo-labels in our method.

To further improve matching accuracy, we introduce a cross-modal feature enhancement module tailored for semi-supervised training. Unlike prior feature enhancement methods that rely on labeled data, our module uses self-attention to suppress modality-specific features and cross-attention to reinforce shared features. It is trained with a mutual independence objective, promoting clean separation between shared and specific features across SAR and optical modalities—without requiring ground-truth labels.

We integrate this enhancement mechanism into our overall multiscale matching framework, resulting in \baby, a semi-supervised pipeline designed for efficient and accurate SAR-optical image matching under limited supervision.

The main contributions of our work are summarized as follows:

\begin{itemize}
\item We propose a multiscale matching method that jointly exploits shallow-level and deep-level features to improve robustness in SAR-optical image matching. The integration of multiscale information reduces mismatching rates compared to using single-level features alone.

\item We design a cross-modal feature enhancement module that utilizes self-attention and cross-attention mechanisms to suppress modality-specific features and reinforce modality-shared features. The module is trained with a mutual independence loss and is compatible with semi-supervised learning.

\item We introduce \baby, a semi-supervised multiscale matching pipeline that leverages pseudo-labeling and multiscale heatmaps. Our method achieves competitive matching accuracy using only a small proportion of labeled data.
\end{itemize}

\section{Related Works}

\subsection{Supervised SAR-optical Image Matching}
Existing supervised SAR-optical image matching methods can be broadly divided into template-based and feature-based approaches. Feature-based matching approaches aim to extract discrete feature descriptors that are invariant across modalities, using dedicated feature extractors~\cite{nie2025promptmidmodalinvariantdescriptors,XU2023103433}. These descriptors are then matched via strategies commonly adopted in non-remote sensing image matching. For example, Zhang et al.~\cite{Zhang2024MultimodalRS} extract features using a U-Net-like backbone and employ a vision transformer to predict the matching position on the optical image based on the SAR image input. Similarly, Li et al.~\cite{10777529} utilize an attention mechanism for denoising, followed by a similarity matrix for matching feature descriptors. Although feature-based methods can achieve high accuracy in areas with distinctive landmarks such as urban regions, they tend to suffer from high mismatching rates when applied to homogeneous terrains, such as agricultural fields, due to their overlooking of geometric correspondence between features.

Template-based methods treat one image in the pair as the template and the other as the floating image~\cite{app13137701,HUGHES2020166}. Feature maps are extracted via neural networks and traditional template-matching metrics such as normalized cross-correlation (NCC) are applied to compute matching positions~\cite{fftmatching}. For example, Fang et al.~\cite{9507635} introduces a Siamese UNet for feature extraction and applies Fast Fourier Transform-accelerated NCC (FFT-NCC) for efficient matching. However, the accuracy of such methods is still undermined by the inherent differences in feature maps extracted from different image modalities. To address this, MARU-Net~\cite{10129005} incorporates attention gates into the U-Net to facilitate feature alignment. CAMM~\cite{CAMM} enhances feature distinctiveness and repeatability by incorporating structural information from both modalities. DCDM~\cite{gou2024interpretable} uses diffusion-based translation to project SAR and optical images into a shared latent space. Although these methods improve performance, their dependence on extensive labeled data remains a major limitation in real-world applications.

\subsection{Linear Attention Mechanism in Vision Tasks}
Attention mechanisms—including self-attention and cross-attention—are widely used in vision tasks for adaptive feature extraction and modality fusion. However, traditional softmax-based attention requires computing the dot product $QK^T$, leading to a quadratic complexity of $O(N^2)$ with respect to sequence length $N$:

\begin{equation}
  \mathrm{Attn}(Q,K,V) = \text{Softmax}(QK^T)V
\end{equation}

In vision applications, where $N = H \times W$ corresponds to the number of pixels, this quadratic complexity becomes a major computational bottleneck. To mitigate this, linear attention mechanisms~\cite{katharopoulos2020transformersrnnsfastautoregressive} replace softmax with a kernel-based similarity: $\phi(Q)\phi(K)^T$, where $\phi(\cdot) = \mathrm{ELU}(\cdot) + 1$. This formulation allows the intermediate matrix $\phi(K)^T V$ to be computed first, reducing the overall time complexity to $O(N)$:

\begin{equation}
  \mathrm{LinearAttn}(Q,K,V) = \phi(Q) (\phi(K)^T V)
\end{equation}
\section{\baby Method}

\begin{figure*}[h]
  \centering
  \includegraphics[width=\textwidth]{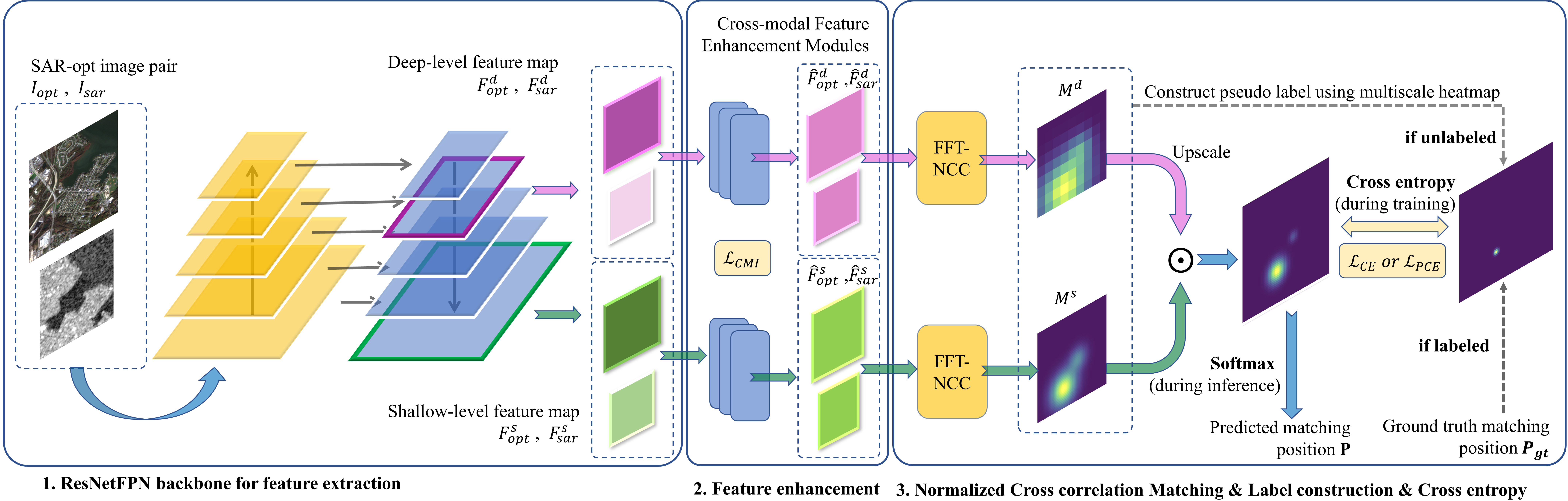}
  \caption{Overview of the Proposed Method}
  \label{fig:overview}
\end{figure*}

The overall framework of \baby is illustrated in Figure~\ref{fig:overview}. 
It comprises two main modules, namely the \emph{Siamese Feature Extraction Backbone} and the \emph{Cross-modal Feature Enhancement Module}. To address the issue of scarce labeled data and exploit the massive unlabeled data, we construct pseudo heatmaps of unlabeled data based on their both deep-level and shallow-level features. The matching model is then trained using both cross-entropy and cross-modal mutual independence losses on labeled and pseudo-labeled data, forming a semi-supervised training pipeline.

Specifically, given an input image $I_{opt}$ with size ($H_{opt}$, $W_{opt}$) as the template image and $I_{SAR}$ with a smaller size of ($H_{SAR}$, $W_{SAR}$) as the floating image, our goal is to predict the correct matching position $P$ for $I_{SAR}$ on $I_{opt}$. For labeled data, the ground truth matching position is denoted as $P_{gt}$. 
We first employ a Siamese ResNet with Feature Pyramid Network (ResNetFPN) backbone, using shared weights for SAR-optical image pairs, to extract deep-level feature maps $F_{opt}^{d}$ and $F_{SAR}^{d}$ with sizes ($H_{opt}/8$, $H_{opt}/8$) and ($H_{SAR}/8$, $W_{SAR}/8$), respectively, as well as shallow-level feature maps $F_{opt}^{s}$ and $F_{SAR}^{s}$ with sizes ($H_{opt}$, $W_{opt}$) and ($H_{SAR}$, $W_{SAR}$). Next, we use two cross-modal feature enhancement modules to reinforce modality-shared information while suppressing modality-specific information in both deep-level and shallow-level feature map pairs. 

Subsequently, we apply Fast Fourier Transform-accelerated Normalized Cross-Correlation (FFT-NCC)~\cite{FNCC}, a widely used method in template matching~\cite{Reddy1996AnFT,buniatyan2017deeplearningimprovestemplate}, to the enhanced multiscale feature map pairs. This yields a shallow-level similarity heatmap $M^{s}$ of size ($H_{opt}$-$H_{SAR}$, $W_{opt}$-$W_{SAR}$), computed from the shallow-level feature maps, and a deep-level similarity heatmap $M^{d}$ of size (($H_{opt}$-$H_{SAR}$)/8, ($W_{opt}$-$W_{SAR}$)/8), computed from the deep-level feature maps.

During inference, we perform a softmax operation on the Hadamard (element-wise) product between the shallow-level and upscaled deep-level similarity heatmap ($upscale(M^{d}) \odot M^{s}$) to predict the matching position $P$.

\paragraph{Multilevel Feature Extraction Backbone}
We employ a Siamese ResNet with a Feature Pyramid Network (ResNetFPN) backbone using shared weights for multi-level feature extraction. To enable pixel-wise accurate matching on shallow-level feature maps, it is essential that their resolution matches that of the original input image pairs. Therefore, unlike conventional feature pyramids~\cite{lin2017featurepyramidnetworksobject}, which extract features at four levels from 1/2 to 1/16 of the original resolution, we omit downsampling in the first convolutional block of the ResNet, and introduce an additional pyramid level by adding a lateral connection that merges the output of the first block into the top-level feature map. From the resulting pyramid, we designate the features at 1/8 resolution level as deep-level feature maps, and the features at original resolution level as shallow-level feature maps.

\paragraph{Semi-supervised Training Pipeline}
We utilize both labeled and unlabeled data during training. Specifically, a cross-entropy loss ($\mathcal{L}_{CE}$) is computed for labeled data, while a pseudo cross-entropy loss ($\mathcal{L}_{PCE}$) is used for unlabeled data. 

For labeled samples, one-hot ground-truth heatmaps $M_{gt}^{d}$ and $M_{gt}^{s}$ are generated based on the known matching positions ($P_{gt}$). The cross-entropy losses calculated between these ground-truth heatmaps and the predicted deep-level ($M^{d}$) and shallow-level ($M^{s}$) heatmaps are summed to form the supervised loss $\mathcal{L}_{CE}$.

For unlabeled data, we upscale the deep-level similarity heatmap $M^{d}$ to match the shallow-level resolution and then produce a pseudo shallow-level heatmap $M_{pseudo}^{s}$ by applying a Hadamard product between the shallow-level ($M^{s}$) and upscaled deep-level ($M^{d}$) heatmaps. The pseudo cross-entropy loss is then computed between the shallow-level prediction $M^{s}$ and the pseudo-label $M_{pseudo}^{s}$.

This pseudo-labeling strategy effectively leverages the complementary strengths of deep-level and shallow-level matching. Deep-level matching converges quickly and yields robust predictions even with limited supervision, thanks to its broader geometric and semantic context, despite its lower spatial resolution (1/8). In contrast, shallow-level matching can achieve pixel-level precision but is more prone to errors under scarce labels. By combining the reliability of deep-level predictions with the precision of shallow-level features, we generate pseudo-labels that help correct mismatches at the shallow level—without the need for extensive manual annotation.

Additionally, a cross-modal feature mutual independence loss $\mathcal{L}_{CMI}$ is applied to both labeled and unlabeled data to train the cross-modal feature enhancement modules for both modalities.

\subsection{Cross-modal Feature Enhancement Module}

\begin{figure}[h]
  \centering
  \includegraphics[width=\linewidth]{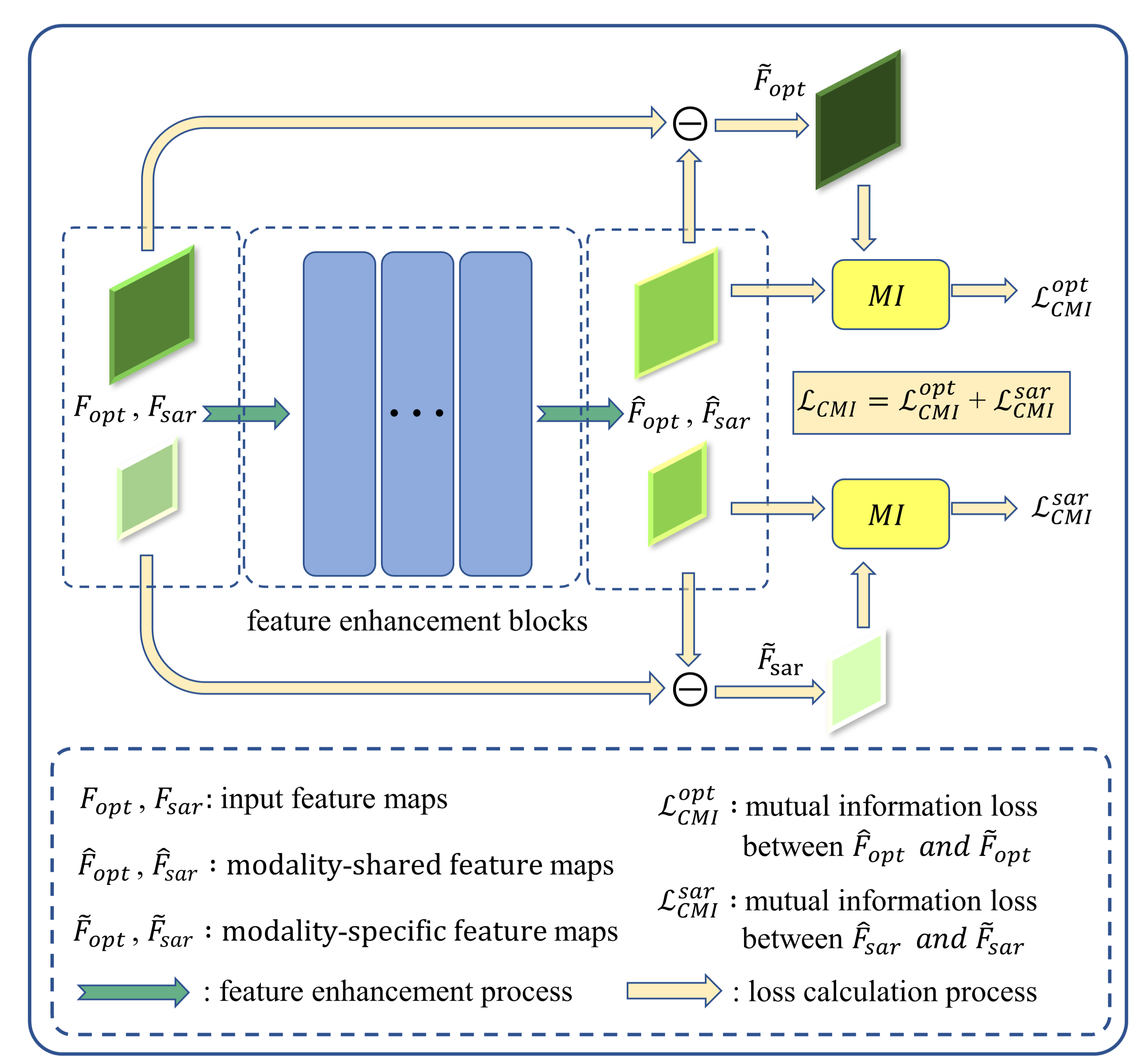}
  \caption{The Overall Structure and Loss Calculation Process of the Proposed Cross-modal Feature Enhancement Module}
  \label{fig:feature_enhance_module}
\end{figure}

Due to inherent modality differences between SAR and optical images, multiscale feature maps extracted by the Siamese backbone inevitably contain modality-specific features. However, traditional feature enhancement methods are limited by the scarcity of labeled data. To mitigate the influence of modality-specific features on matching accuracy, we introduce a cross-modal feature enhancement module. This module employs attention mechanisms to suppress modality-specific information and strengthen modality-shared features across both shallow-level and deep-level feature maps. Our cross-modal feature enhancement module consists of one or multiple feature enhancement blocks stacked sequentially. Figure~\ref{fig:feature_enhance_module} illustrates the overall structure of this module.

\subsubsection{Feature Enhancement Block}
\begin{figure}[h]
  \centering
  \includegraphics[width=\linewidth]{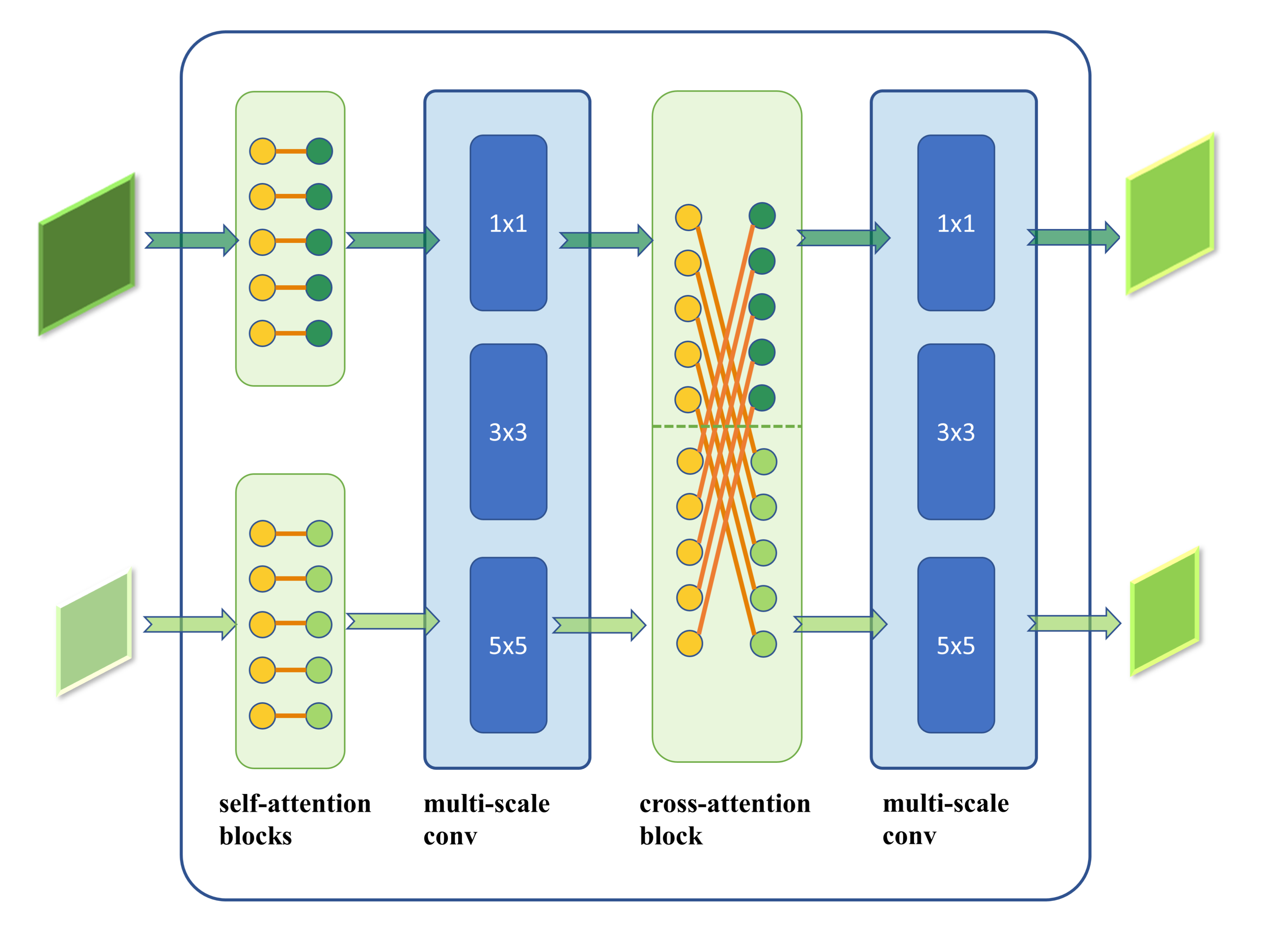}
  \caption{The Feature Enhancement Block in Cross-modal Feature Enhancement Module}
\label{fig:feature_enhance_block}
\end{figure}

 Figure~\ref{fig:feature_enhance_block} illustrates the architecture of a single feature enhancement block in the cross-modal feature enhancement module. Each block contains two separate self-attention blocks, one for each modality, and a shared cross-modality attention block. Each attention block is followed by a multiscale convolution block to ensure feature consistency. By applying this module to the extracted multiscale feature maps, we aim to suppress features that are unique to either SAR or optical modalities and enhance features that are common to both modalities.

\subsubsection{Attention Block}
\begin{figure}[h]
  \centering
  \includegraphics[width=\linewidth]{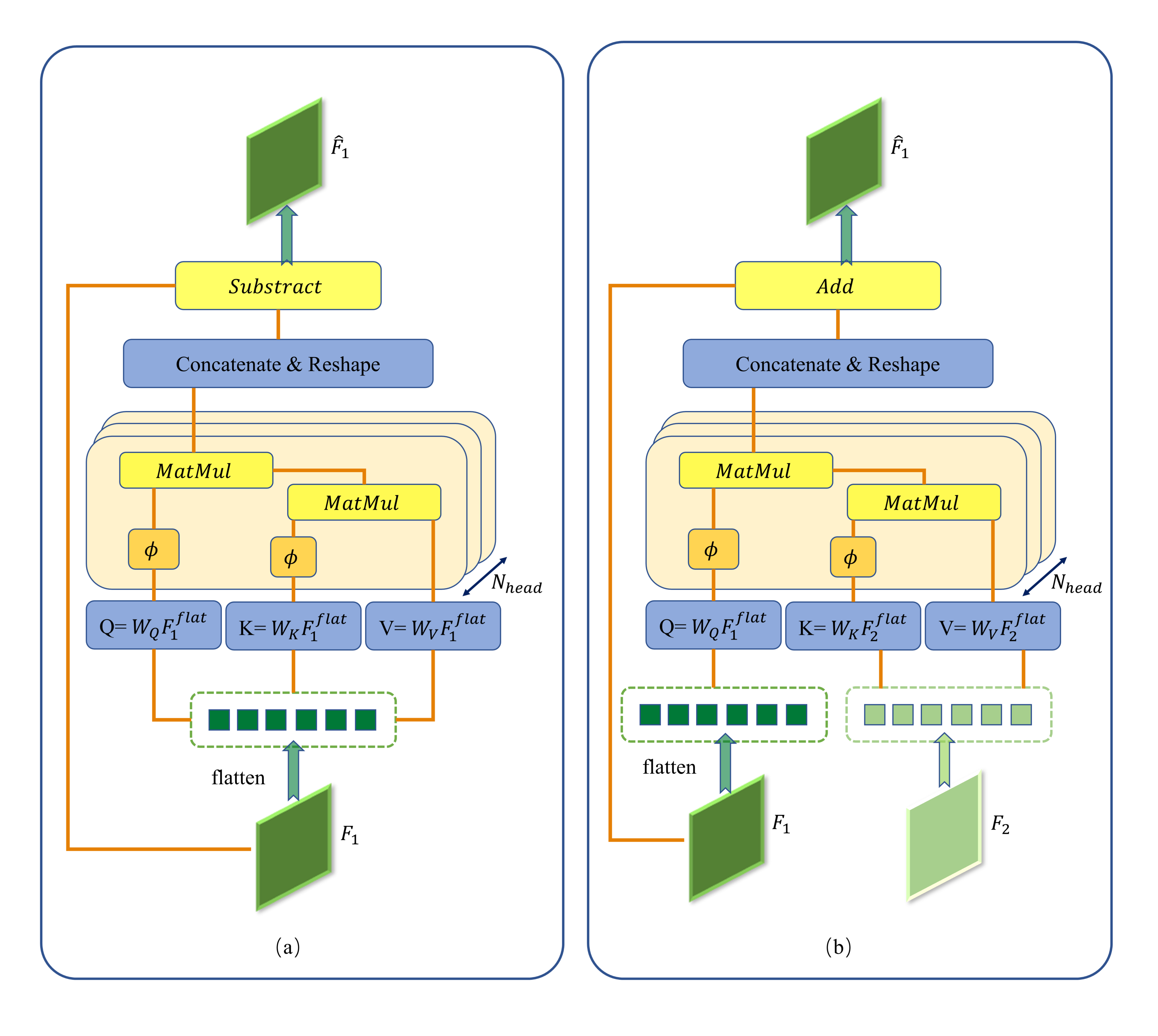}
  \caption{The Linear Attention Block in Cross-modal Feature Enhancement Module. (a) Self-attention Block (b) Cross-attention Block}
  \label{fig:attention_block}
\end{figure}

Figure~\ref{fig:attention_block} illustrates the structure of the self-attention and cross-attention blocks. Both blocks employ a multi-head linear attention mechanism to compute attention.

In the self-attention block, the input feature map $F_1$ s first flattened into a sequence with spatial dimensions $H \times W$, resulting in $F_1^{flat}$. This sequence is then used to compute the query $Q$, key $K$ and value $V$ projections via three learnable linear layers. The same sequence serves as input for both $Q$, $K$ and $V$ in order to extract modality-specific features. The resulting features are reshaped to their original spatial dimensions, denoted as $F_1^{self}$. Ultimately, we decompose the extracted features from the input feature map by subtracting $F_1^{self}$ from $F_1$. 

In the cross-attention block, two feature maps ($F_1$, $F_2$) from different modalities are processed. After flattening and projection, each sequence serves as the key $K$ and value $V$ for the other, enabling the extraction of modality-shared features through cross-attention. The reshaped features $F_1^{cross}$ and $F_2^{cross}$ are then added to their respective input feature maps $F_1$ and $F_2$.  Following the attention operation, a multiscale convolution block is applied to maintain spatial consistency.

\subsubsection{Loss for Cross-modal Feature Enhancement Module}

 By applying the cross-modal feature enhancement module, we aim to discard modality-specific features from the feature map pairs and retain only the enhanced modality-shared features for matching. To this end, the interdependence between modality-shared and modality-specific features should be minimized. Specifically, we decouple the modality-specific feature map $\widetilde{F}$ from the original feature map by subtracting the extracted modality-shared feature map $\widehat{F}$. Mutual information is then employed as the objective function to quantify and minimize the dependency between these two types of features. The discrete mutual information is mathematically formulated as follows:
\begin{equation}
    I(X,Y) = \sum_{x \in \mathcal{X}}\sum_{y \in \mathcal{Y}} p(x,y)\log\frac{p(x,y)}{p(x)p(y)}
\end{equation}
The goal of this calculation is to minimize the mutual information (MI) between two feature representations, thereby promoting their independence. This principle is grounded in the idea that the retained modality-shared features and discarded modality-specific features should inherently exhibit orthogonality and minimal overlap. The marginal probability distributions $p(x)$ and $p(x)$, along with the joint probability distribution $p(x,y)$, are computed from the flattened modality-shared and modality-specific feature maps, denoted as $\widehat{F}^{flat}$ and $\widetilde{F}^{flat}$, respectively.The total cross-modal feature mutual independence loss $\mathcal{L}_{CMI}$ is composed of two components: the mutual independence loss for the SAR modality $\mathcal{L}_{CMI}^{sar}$, and the mutual independence loss for the optical modality $\mathcal{L}_{CMI}^{opt}$. This loss calculation requires no supervision.

\begin{equation}
    \begin{split}
        \mathcal{L}_{CMI}^{sar} = I(\widehat{F}_{sar}^{flat},\widetilde{F}_{sar}^{flat}) \\
        \mathcal{L}_{CMI}^{opt} = I(\widehat{F}_{opt}^{flat},\widetilde{F}_{opt}^{flat}) \\
        \mathcal{L}_{CMI} = \mathcal{L}_{CMI}^{sar} + \mathcal{L}_{CMI}^{opt}
    \end{split}
\end{equation}

\subsection{Loss Functions}

To train our \baby model in a semi-supervised manner, we define a supervised loss $\mathcal{L}_{sup}$ for labeled data and an unsupervised loss $\mathcal{L}_{unsup}$ for unlabeled data. The model is optimized using the combined loss $\mathcal{L}_{sup} + \mathcal{L}_{unsup}$, leveraging both labeled and unlabeled samples simultaneously.

\begin{equation}
    \mathcal{L}_{semi}=\mathcal{L}_{sup}(labeled)+\mathcal{L}_{unsup}(all)
\end{equation}

\subsubsection{Loss for labeled data}
The supervised loss $\mathcal{L}_{sup}$ is composed of a linear combination of cross-entropy loss $\mathcal{L}_{CE}$ and cross-modal feature mutual independence loss $\mathcal{L}_{CMI}$. First, ground truth heatmaps $M_{gt}^{d}$ and $M_{gt}^{s}$ are generated for the deep-level and shallow-level predictions based on the ground truth matching position $P_{gt}$. These heatmaps are hard labels, where only the ground truth position is marked as 1 and all others as 0. We compute the deep-level matching loss $\mathcal{L}_{CE}^{d}$ as the cross-entropy between $M^{d}$ and $M_{gt}^{d}$, and the shallow-level matching loss as deep-level matching loss $\mathcal{L}_{CE}^{s}$ as the cross-entropy between $M^{s}$ and $M_{gt}^{s}$.

\begin{equation}
    \begin{split}
        \mathcal{L}_{CE}=\mathcal{L}_{CE}^{d}+\mathcal{L}_{CE}^{s}\\
        \mathcal{L}_{CMI}=\mathcal{L}_{CMI}^{d}+\mathcal{L}_{CMI}^{s}\\
        \mathcal{L}_{sup}=\mathcal{L}_{CE}+\mathcal{L}_{CMI}
    \end{split}
\end{equation}

\subsubsection{Loss for unlabeled data}

 The unsupervised loss $\mathcal{L}_{unsup}$ is composed of a linear combination of pseudo cross-entropy loss $\mathcal{L}_{PCE}$ and cross-modal feature mutual independence loss $\mathcal{L}_{CMI}$. To compute the pseudo-label supervision, we first generate a pseudo ground truth heatmap $M_{pgt}^{s}$ by combining the upscaled deep-level similarity heatmap $M^{d}$ and the shallow-level heatmap $M^{s}$ through element-wise multiplication. The pseudo cross-entropy loss $\mathcal{L}_{PCE}$ is then calculated between the shallow-level prediction $M^{s}$ and the pseudo ground truth heatmap $M_{pgt}^{s}$. 
 The cross-modal mutual independence loss $\mathcal{L}_{CMI}$ loss is also applied to further train the feature enhancement module.

\begin{equation}
    \begin{split}
        M_{pgt} = upscale(M^{d}) \odot M^{s}
        \\
        \mathcal{L}_{PCE} = CE(M^{s}, M_{pgt}^{s}) \\ \mathcal{L}_{unsup}=\mathcal{L}_{PCE}+\mathcal{L}_{CMI}
    \end{split}
\end{equation}

\section{Experiments}
To evaluate the performance of our \baby method, we train and test our model on two different datasets (SEN1-2 and QXS-SAROPT), and compare its matching accuracy and computational efficiency with one existing semi-supervised SAR-optical image matching method (Semi-I2I) and three fully supervised methods (FFT+U-Net, MARU-Net, OSMNet).
\subsection{Experiments Settings}
\subsubsection{Datasets}
\paragraph{SEN1-2 Dataset}
The SEN1-2 dataset~\cite{schmitt2018sen12datasetdeeplearning} is a large-scale collection designed to advance deep learning research in synthetic aperture radar (SAR) and optical data fusion. It comprises 282,384 pairs of corresponding image patches from the Sentinel-1 and Sentinel-2 satellites, covering diverse global locations across all seasons. Each patch measures 256×256 pixels with a spatial resolution of 10 meters. After shuffling the dataset using a fixed random seed, we take the first 90.7\% (256,000 image pairs) as the training set and the remaining 9.3\% (26,384 image  pairs) as the testing set.

\paragraph{QXS-SAROPT Dataset}
The QXS-SAROPT dataset~\cite{huang2021qxssaroptdatasetdeeplearning} contains 20,000 pairs of SAR-optical image patches. The SAR image patches are obtained from SAR satellite GaoFen-3 images, and the corresponding optical patches are sourced from Google Earth. These images cover three port cities: San Diego, Shanghai and Qingdao. All images have a fixed size of 256×256 pixels. In our experiment, we shuffle the dataset using a fixed random seed and use the first 80\% (16,000 image pairs) as the training set and the remaining 20\% (4,000 image  pairs) as the testing set.

\subsubsection{Baseline Methods}
\paragraph{Semi-supervised Baseline}
We adopt the Semi-I2I method~\cite{Du2022ASI} as the semi-supervised baseline. Since the available source code provides only an SAR-to-optical translation framework, we first translate SAR images into the optical modality using Semi-I2I, and subsequently perform template matching on the resulting image pairs using FFT-NCC.

\paragraph{Supervised Baselines}
To further evaluate the performance of our method, we adopt several state-of-the-art supervised matching methods, namely FFT+U-Net~\cite{9507635}, MARU-Net~\cite{Du2022ASI}, OSMNet~\cite{9609993} and DCDM~\cite{gou2024interpretable} as our baselines. For MARU-Net and OSMNet, we directly use the publicly available implementations from their original papers. For FFT+U-Net, we reimplement the method in PyTorch, replicating the network structure and hyperparameters used in the original publication.

\subsubsection{Implementation Details}
\paragraph{Environment}
Our method is implemented in PyTorch 2.0.0 with Python 3.8.10 and CUDA 11.8. All experiments, including the training and evaluation of both our method and the baselines, are conducted on an Ubuntu server equipped with an NVIDIA A100 40GB Tensor Core GPU and an Intel Xeon Gold 6230R CPU.

\paragraph{Hyperparameters Settings}
During the semi-supervised training process, we set the batch size to 16 and use an AdamW optimizer with a weight decay of $1\times 10^{-2}$. To balance accuracy and efficiency, we use only one feature enhancement block in each cross-modal feature enhancement module unless otherwise specified. In all of our experiments—both for our method and the baseline comparisons—the template image size is set to 192 × 192 pixels, and the reference image size to 256 × 256 pixels. All the modules are initialized randomly.

Unless otherwise specified, the ratio of labeled to unlabeled batches is 1:15 (i.e., 6.25\% of batches contain labeled data), with one labeled and fifteen unlabeled batches per iteration. For the SEN1-2 dataset, we train the model for 5 epochs with an initial learning rate of $5\times 10^{-5}$. For the QXS-SAROPT dataset, due to its lower data volume, we train the model for 10 epochs using a higher learning rate of $5\times10^{-4}$ to ensure convergence.

\subsubsection{Evaluation Metrics}
In our experiments, we adopt four commonly used metrics for remote sensing image (RSI) matching tasks: RMSE, CMR(T=1), CMR(T=5), and RMSE(T=5). RMSE measures the average root mean square error (i.e., Euclidean distance) between the predicted and ground truth matching positions across all test samples. CMR(T=$\alpha$) denotes the correct matching rate under a threshold $\alpha$, where predictions within $\alpha$ pixels of the ground truth are considered correct matches. RMSE(T=$\alpha$) measures the average root mean square error among all predictions that fall within the 
$\alpha$-pixel threshold. Inference efficiency is evaluated based on the average processing time (in milliseconds) per image pair.

\subsection{Main Results}
\subsubsection{Quantitative Analysis}

Table~\ref{tab:sen1-2result} presents the matching performance of our method and baseline methods on the SEN1-2 dataset. The results demonstrate that our method, trained with only 6.25\% labeled data, outperforms both supervised and semi-supervised baselines in terms of CMR (T=5) and RMSE. Remarkably, our approach remains highly competitive even against fully supervised methods trained with 100\% labeled data. These outcomes align with our goal of achieving robust matching performance by effectively leveraging unlabeled data. 

\begin{table}[h]
\setlength{\tabcolsep}{4mm}
\centering
\setlength{\abovecaptionskip}{0.5ex}
\begin{threeparttable}
\caption{The Matching Results on the SEN1-2 Dataset}
\label{tab:sen1-2result}
\begin{tabular}{lccccc}
\toprule
\textbf{Methods} & 
\textbf{CMR(T=1)} & 
\textbf{CMR(T=5)} & 
\textbf{RMSE(T=5)} & 
\textbf{RMSE(All)} &
\textbf{Time(ms)} \\
\midrule
FFT+U-Net\tnote{*}  & 0.4239 & 0.8250 & 1.7226 & 6.8715 & \textbf{26.3}\\
MARU-Net\tnote{*}   & \textbf{0.4855} & 0.8943 & \textbf{1.4106} & 5.0945 & 29.8\\
OSMNet\tnote{*}     & 0.4501 & 0.9175 & 1.6309 & 4.5033 & 42.2\\
Semi-I2I\tnote{**}  & 0.2927 & 0.7613 & 2.3981 & 8.1639 & 161.4\\
Ours\tnote{**}      & 0.4480 & \textbf{0.9250} & 1.4740 & \textbf{3.7844} & 62.7\\
\bottomrule
\\
\end{tabular}
\end{threeparttable}

\begin{threeparttable}
\caption{The Matching Results on the QXS-SAROPT Dataset}
\label{tab:QXSresult}
\begin{tabular}{lccccc}
\toprule
\textbf{Methods} & 
\textbf{CMR(T=1)} & 
\textbf{CMR(T=5)} & 
\textbf{RMSE(T=5)} & 
\textbf{RMSE(All)} &
\textbf{Time(ms)} \\
\midrule
FFT+U-Net\tnote{*}  & 0.2436 & 0.6947 & 2.8783 & 7.6092 & \textbf{26.1} \\
MARU-Net\tnote{*}   & 0.2918 & 0.8004 & 2.5629 &  5.9390 & 30.5\\
OSMNet\tnote{*}     & 0.2940 & \textbf{0.8498} & 2.4183 & \textbf{5.6268} & 41.9\\
Semi-I2I\tnote{**}  & 0.1957 & 0.6406 & 3.6143 &  10.6475 & 158.0\\
Ours\tnote{**}      & \textbf{0.3275} & 0.8248 & \textbf{2.2665} &  5.7239 & 61.9 \\
\bottomrule
\end{tabular}
 \begin{tablenotes}
    \footnotesize
    \item[*] Fully Supervised Methods, trained with 100\% labeled data
    \item[**] Semi-supervised Methods, trained with 6.25\% labeled data and 93.75\% unlabeled data
\end{tablenotes}
\end{threeparttable}
\end{table}

Table \ref{tab:QXSresult} shows the results on the QXS-SAROPT dataset. The limited size of the QXS-SAROPT dataset results in performance degradation across both supervised and semi-supervised methods. However, our method achieves superior performance in terms of CMR(T=1) and RMSE(T=5), demonstrating that it maintains reliable performance even under conditions of extreme scarcity of labeled data.

\subsection{Analysis on Pseudo-label Quality}
\begin{figure}[h]
  \centering
  \includegraphics[width=0.7\linewidth]{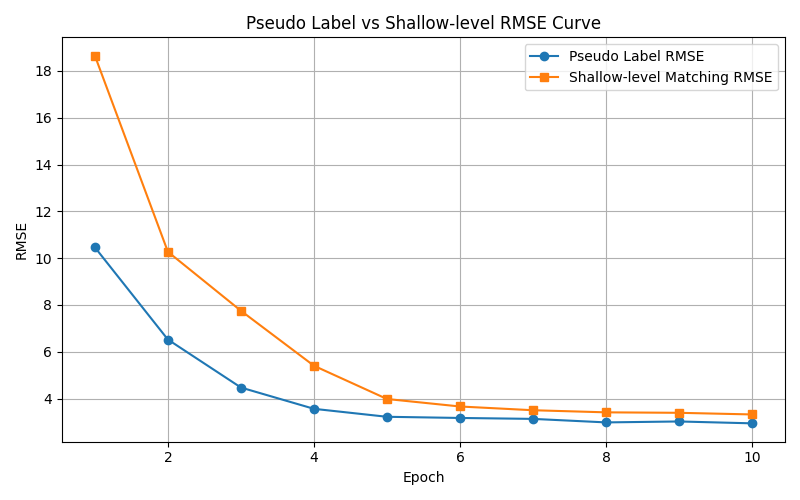}
  \caption{The RMSE Curve of Pseudo Labels and Shallow-level Matching Results on Validation Set During Training}
  \label{fig:loss_curve}
\end{figure}

We evaluated the quality of pseudo-labels during semi-supervised training by comparing their RMSE against shallow-level predictions. The RMSE trend on validation set is shown in Figure~\ref{fig:loss_curve}.

In the early training epochs, pseudo-labels exhibit significantly higher accuracy than shallow-level predictions. This supports our hypothesis that leveraging the robustness of deep-level matching enables reliable pseudo-labels even with limited labeled data. As training progresses, both pseudo-label and shallow-level accuracies improve steadily, validating the effectiveness of our refinement strategy. By convergence, the gap between pseudo-label and shallow-level predictions narrows, indicating that shallow-level matching benefits from consistent pseudo-label guidance.

\begin{figure}[h]
  \centering
  \includegraphics[width=0.7\linewidth]{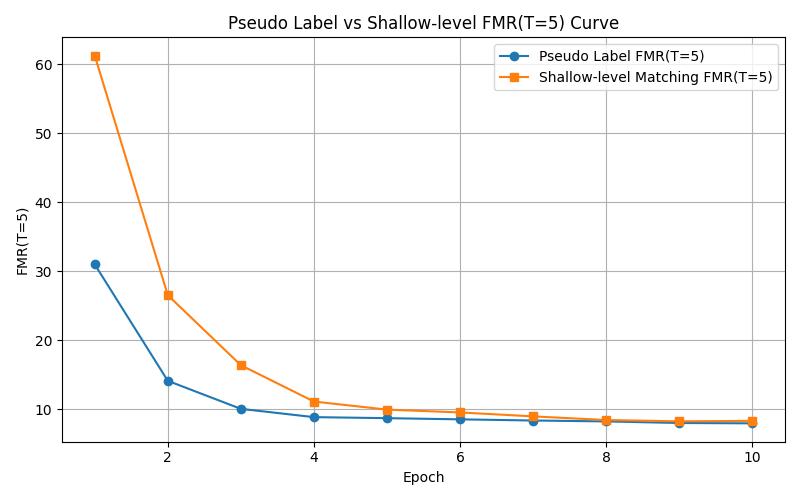}
  \caption{The FMR(T=5) Curve of Pseudo Labels and Shallow-level Matching Results on Validation Set During Training}
  \label{fig:fmr_curve}
\end{figure}

Additionally, we evaluated the false matching rate at a 5-pixel threshold, denoted as FMR(T=5), where any prediction deviating more than 5 pixels from the ground truth is considered a false match. As shown in Figure~\ref{fig:fmr_curve}, our pseudo-labels consistently yield a lower false matching rate than shallow-level predictions, particularly during the early training epochs. This further validates the robustness of our multiscale matching and pseudo-labeling strategy when labeled data is scarce.

\subsection{Influence of Labeled Data Quantity}
\begin{figure}[h]
  \centering
  \includegraphics[width=0.7\linewidth]{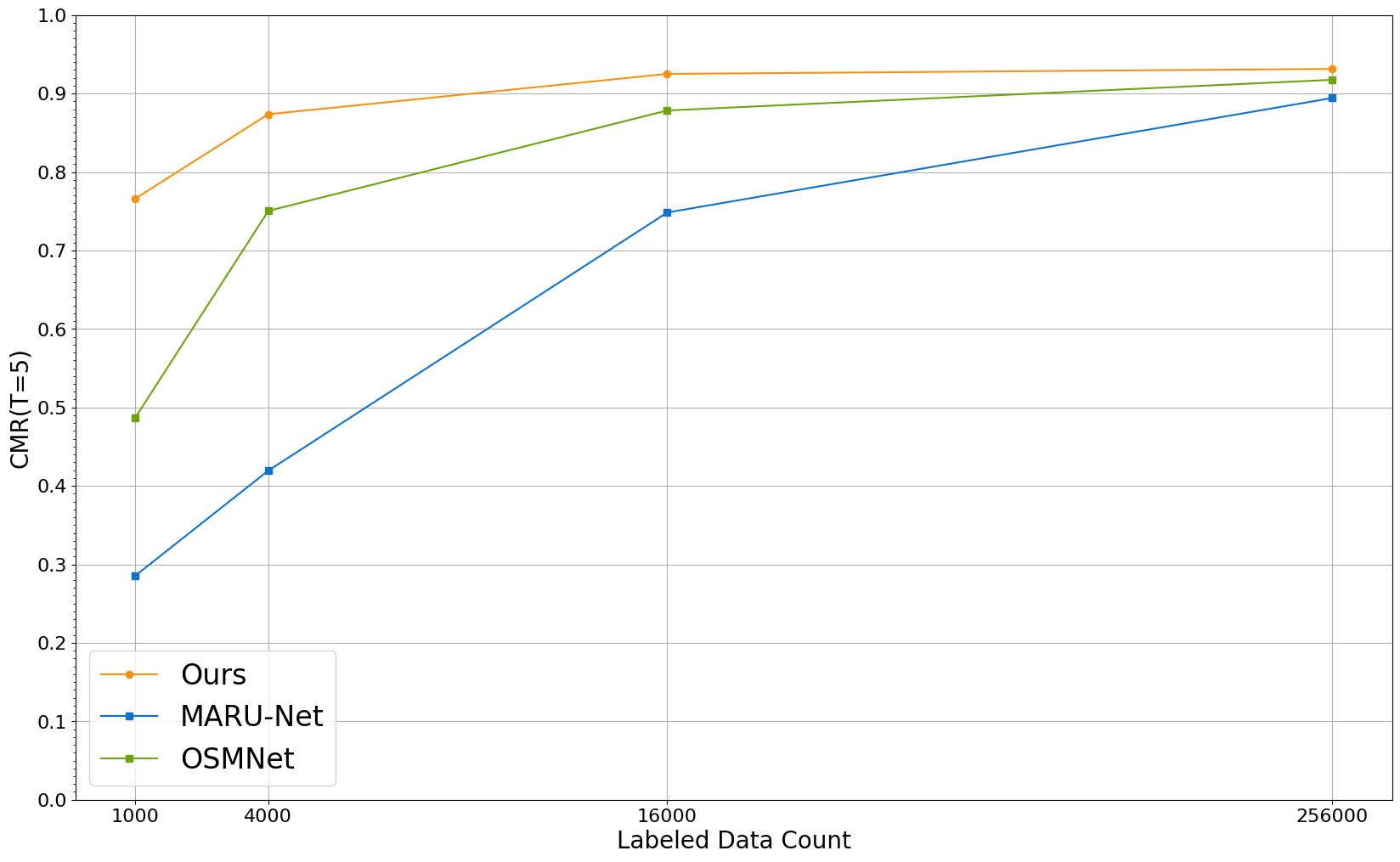}
  \caption{The Influence of Labeled Data Quantity in the SEN1-2 Dataset}
  \label{fig:data_scarcity}
\end{figure}

To examine the impact of labeled data quantity on matching accuracy, we train our model on the SEN1-2 dataset using various labeled-to-unlabeled batch ratios, while keeping all other settings unchanged. For comparison, we include MARU-Net and OSMNet, which demonstrate leading performance among all other tested methods. Since MARU-Net and OSMNet are fully supervised methods, only the labeled portion of the data is used during their training. The results, evaluated using CMR(T=5) and presented in Figure~\ref{fig:data_scarcity}, show that our model maintains satisfactory accuracy even with a minimal amount of labeled data, and the performance gradually improves as more labeled data are introduced. These findings suggest that our method may be a preferable alternative to fully supervised approaches when labeled data are scarce.

\subsubsection{Visualizaion of the NCC Heatmap}

\begin{figure}[h]
  \centering
  \includegraphics[width=0.9\linewidth]{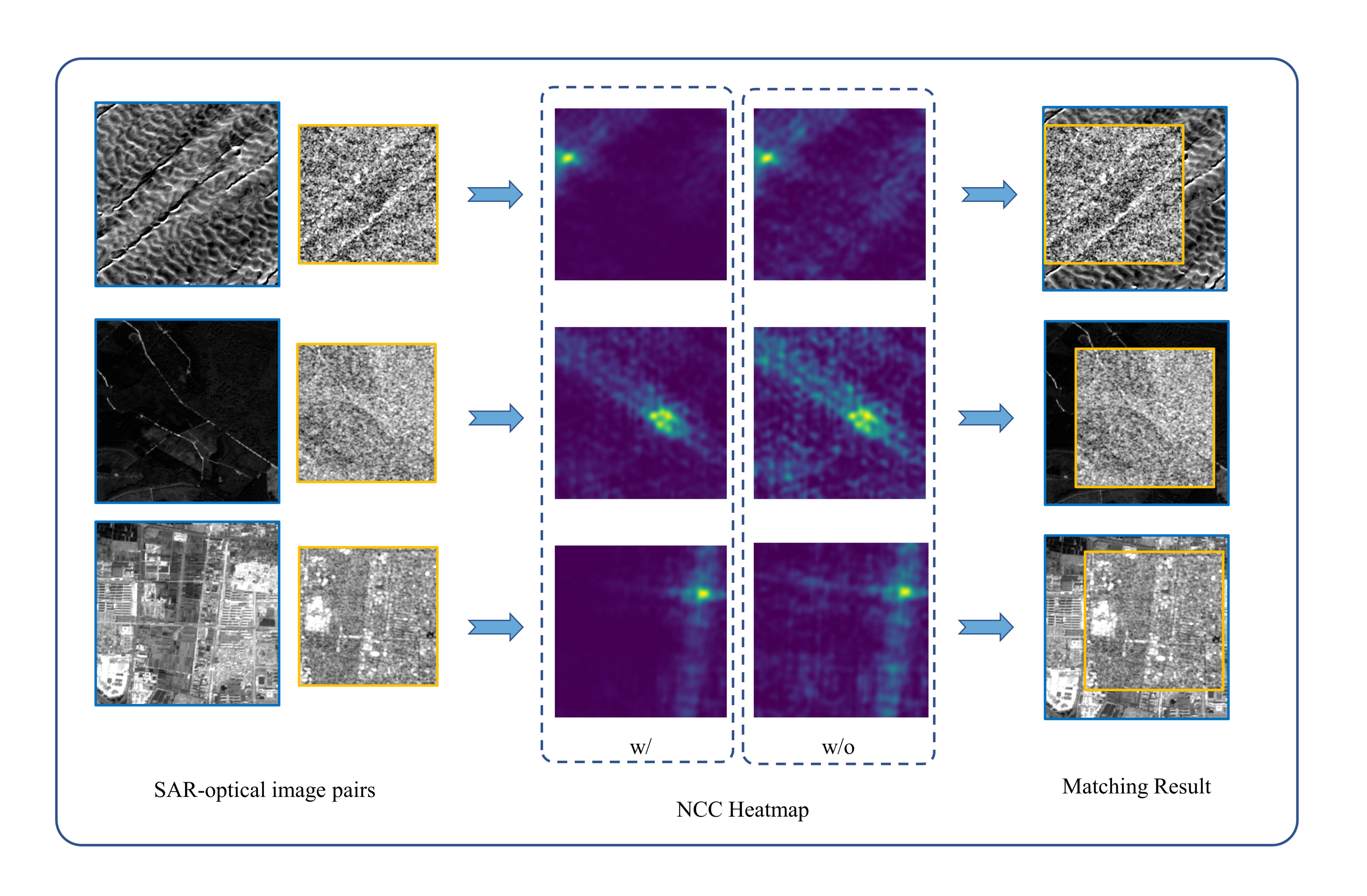}
  \caption{The Visualization of NCC Heatmap with and without Multiscale Matching}
  \label{fig:visualization}
\end{figure}

To demonstrate the effectiveness of our multiscale matching approach, we visualize the NCC heatmaps with and without multiscale matching in Figure \ref{fig:visualization}. From the highlighted regions, we observe that compared to the heatmap generated without multiscale matching, the heatmap from our method exhibits a more concentrated high-similarity region and lower responses in non-matching areas. This aligns with our expectation that multiscale matching enhances the robustness of correspondence estimation.

\subsection{Ablation Study}
To further validate the effectiveness of each proposed component, we perform an ablation study on the SEN1-2 dataset. We begin by decomposing our full model into four components: the ResNetFPN backbone combined with FFT-NCC (denoted as B), multiscale feature matching (denoted as M), the use of unlabeled data in semi-supervised training (denoted as S), and the cross-modal feature enhancement module with $n$ feature enhancement blocks (denoted as $F^n$). We then evaluate the matching performance by progressively adding each component to the backbone. The experimental results are presented in Table \ref{tab:ablation}.

\begin{table}[h]
\setlength{\tabcolsep}{4mm}
\centering
\begin{threeparttable}
\caption{The Results of Ablation Study on the SEN1-2 Dataset}
\label{tab:ablation}
\begin{tabular}{lccccc}
\toprule
\textbf{Methods} & 
\textbf{CMR(T=1)} & 
\textbf{CMR(T=5)} & 
\textbf{RMSE(T=5)} & 
\textbf{RMSE(All)} &
\textbf{Time(ms)} \\
\midrule
B & 0.3307 & 0.8074 & 3.3614 & 8.0382 & 33.1\\
B+M & 0.3152 & 0.8395 & 3.2533 & 7.8141 & 45.9\\
B+M+S & 0.3659 & 0.8816 & 2.8320 & 4.5748 & 45.7\\
B+M+S+$F^1$ & 0.4480 & 0.9250 & 1.4740 & 3.7844 & 62.7 \\
B+M+S+$F^3$ & 0.4701 & 0.9286 & 1.3692 & 3.6990 & 99.5 \\
\bottomrule
\end{tabular}
\end{threeparttable}
\end{table}

The results of the ablation study indicate that incorporating unlabeled data in the semi-supervised training process (S) plays a crucial role in stabilizing the RMSE and enhancing matching accuracy. Although the multiscale matching component (M) does not directly improve accuracy, it is essential for pseudo-label construction and therefore remains a necessary part of our framework. Regarding the cross-modal feature enhancement module, even a single feature enhancement block ($F^1$) substantially improves overall performance. While increasing the number of enhancement blocks to three ($F^3$) further boosts the model’s ability to perform precise matching (particularly in terms of (CMR with T=1), it also significantly increases computational cost.
\section{Conclusion}
In this paper, we introduced \baby, a semi-supervised multiscale matching method designed to address the challenge of SAR-optical image matching under limited labeled data. Our approach leverages the complementary nature of shallow and deep-level features, combining labeled data with abundant unlabeled pairs via pseudo-labeling and multiscale similarity heatmaps. A cross-modal feature enhancement module based on self- and cross-attention further improves performance by emphasizing modality-shared features. Experiments on the SEN1-2 and QXS-SAROPT datasets confirmed our model's competitive performance relative to both supervised and semi-supervised baselines, and ablation studies verified the effectiveness of our key components. Future work will explore advanced unsupervised feature extraction methods to improve generalization in scenarios with extremely limited labeled data.

\bibliographystyle{unsrt}  
\bibliography{references}

\begin{thebibliography}{10}

\bibitem{rs15030850}
Oscar Sommervold, Michele Gazzea, and Reza Arghandeh.
\newblock A survey on sar and optical satellite image registration.
\newblock {\em Remote Sensing}, 15(3), 2023.

\bibitem{zhang2025multiresolutionsaropticalremote}
Wenfei Zhang, Ruipeng Zhao, Yongxiang Yao, Yi~Wan, Peihao Wu, Jiayuan Li, Yansheng Li, and Yongjun Zhang.
\newblock Multi-resolution sar and optical remote sensing image registration methods: A review, datasets, and future perspectives, 2025.

\bibitem{1344163}
J.~Inglada and A.~Giros.
\newblock On the possibility of automatic multisensor image registration.
\newblock {\em IEEE Transactions on Geoscience and Remote Sensing}, 42(10):2104--2120, 2004.

\bibitem{Ayubi:24}
Gast\'{o}n~A. Ayubi, Bartlomiej Kowalski, and Alfredo Dubra.
\newblock Normalized weighted cross correlation for multi-channel image registration.
\newblock {\em Opt. Continuum}, 3(5):649--665, May 2024.

\bibitem{ye2021improvingcoregistrationsentinel1sar}
Yuanxin Ye, Chao Yang, Bai Zhu, Youquan He, and Huarong Jia.
\newblock Improving co-registration for sentinel-1 sar and sentinel-2 optical images, 2021.

\bibitem{Lowe2004DistinctiveIF}
David~G. Lowe.
\newblock Distinctive image features from scale-invariant keypoints.
\newblock {\em International Journal of Computer Vision}, 60:91--110, 2004.

\bibitem{6824220}
Flora Dellinger, Julie Delon, Yann Gousseau, Julien Michel, and Florence Tupin.
\newblock Sar-sift: A sift-like algorithm for sar images.
\newblock {\em IEEE Transactions on Geoscience and Remote Sensing}, 53(1):453--466, 2015.

\bibitem{8272317}
Yuming Xiang, Feng Wang, and Hongjian You.
\newblock Os-sift: A robust sift-like algorithm for high-resolution optical-to-sar image registration in suburban areas.
\newblock {\em IEEE Transactions on Geoscience and Remote Sensing}, 56(6):3078--3090, 2018.

\bibitem{10.1007/s11263-020-01359-2}
Jiayi Ma, Xingyu Jiang, Aoxiang Fan, Junjun Jiang, and Junchi Yan.
\newblock Image matching from handcrafted to deep features: A survey.
\newblock {\em Int. J. Comput. Vision}, 129(1):23–79, January 2021.

\bibitem{Xu_2024}
Shibiao Xu, Shunpeng Chen, Rongtao Xu, Changwei Wang, Peng Lu, and Li~Guo.
\newblock Local feature matching using deep learning: A survey.
\newblock {\em Information Fusion}, 107:102344, July 2024.

\bibitem{ye20243mosmultisourcesmultiresolutionsmultiscenes}
Yibin Ye, Xichao Teng, Shuo Chen, Yijie Bian, Tao Tan, and Zhang Li.
\newblock 3mos: Multi-sources, multi-resolutions, and multi-scenes dataset for optical-sar image matching, 2024.

\bibitem{8898635}
Lloyd~Haydn Hughes, Nina Merkle, Tatjana Bürgmann, Stefan Auer, and Michael Schmitt.
\newblock Deep learning for sar-optical image matching.
\newblock In {\em IGARSS 2019 - 2019 IEEE International Geoscience and Remote Sensing Symposium}, pages 4877--4880, 2019.

\bibitem{schmitt2018sen12datasetdeeplearning}
Michael Schmitt, Lloyd~Haydn Hughes, and Xiao~Xiang Zhu.
\newblock The sen1-2 dataset for deep learning in sar-optical data fusion, 2018.

\bibitem{hughes2019semi}
LH~Hughes and M~Schmitt.
\newblock A semi-supervised approach to sar-optical image matching.
\newblock {\em ISPRS Annals of the Photogrammetry, Remote Sensing and Spatial Information Sciences}, 4:71--78, 2019.

\bibitem{Du2022ASI}
Wenliang Du, Yong Zhou, Hancheng Zhu, Jiaqi Zhao, Zhiwen Shao, and Xiaolin Tian.
\newblock A semi-supervised image-to-image translation framework for sar–optical image matching.
\newblock {\em IEEE Geoscience and Remote Sensing Letters}, 19:1--5, 2022.

\bibitem{Khurshid_2020}
Numan Khurshid, Mohbat Tharani, Murtaza Taj, and Faisal~Z. Qureshi.
\newblock A residual-dyad encoder discriminator network for remote sensing image matching.
\newblock {\em IEEE Transactions on Geoscience and Remote Sensing}, 58(3):2001–2014, March 2020.

\bibitem{nie2025promptmidmodalinvariantdescriptors}
Han Nie, Bin Luo, Jun Liu, Zhitao Fu, Huan Zhou, Shuo Zhang, and Weixing Liu.
\newblock Promptmid: Modal invariant descriptors based on diffusion and vision foundation models for optical-sar image matching, 2025.

\bibitem{XU2023103433}
Wangyi Xu, Xinhui Yuan, Qingwu Hu, and Jiayuan Li.
\newblock Sar-optical feature matching: A large-scale patch dataset and a deep local descriptor.
\newblock {\em International Journal of Applied Earth Observation and Geoinformation}, 122:103433, 2023.

\bibitem{Zhang2024MultimodalRS}
Yong-Xu Zhang, Chaozhen Lan, Haiming Zhang, Guorui Ma, and Heng Li.
\newblock Multimodal remote sensing image matching via learning features and attention mechanism.
\newblock {\em IEEE Transactions on Geoscience and Remote Sensing}, 62:1--20, 2024.

\bibitem{10777529}
Yuan Li, Chuanfeng Wei, Dapeng Wu, Yaping Cui, Peng He, Yuan Zhang, and Ruyan Wang.
\newblock A robust multisource remote sensing image matching method utilizing attention and feature enhancement against noise interference.
\newblock {\em IEEE Transactions on Geoscience and Remote Sensing}, 62:1--21, 2024.

\bibitem{app13137701}
Songlai Han, Xuesong Liu, Jing Dong, and Haiqiao Liu.
\newblock Remote sensing multimodal image matching based on structure feature and learnable matching network.
\newblock {\em Applied Sciences}, 13(13), 2023.

\bibitem{HUGHES2020166}
Lloyd~Haydn Hughes, Diego Marcos, Sylvain Lobry, Devis Tuia, and Michael Schmitt.
\newblock A deep learning framework for matching of sar and optical imagery.
\newblock {\em ISPRS Journal of Photogrammetry and Remote Sensing}, 169:166--179, 2020.

\bibitem{fftmatching}
Kai Briechle and Uwe Hanebeck.
\newblock Template matching using fast normalized cross correlation.
\newblock {\em Proceedings of SPIE - The International Society for Optical Engineering}, 4387, 03 2001.

\bibitem{9507635}
Yuyuan Fang, Jun Hu, Chuan Du, Zhibo Liu, and Lei Zhang.
\newblock Sar-optical image matching by integrating siamese u-net with fft correlation.
\newblock {\em IEEE Geoscience and Remote Sensing Letters}, 19:1--5, 2022.

\bibitem{10129005}
Michele Gazzea, Oscar Sommervold, and Reza Arghandeh.
\newblock Maru-net: Multiscale attention gated residual u-net with contrastive loss for sar-optical image matching.
\newblock {\em IEEE Journal of Selected Topics in Applied Earth Observations and Remote Sensing}, 16:4891--4899, 2023.

\bibitem{CAMM}
Jiaxing Chen, Hongtu Xie, Lin Zhang, Jun Hu, Hejun Jiang, and Guoqian Wang.
\newblock Sar and optical image registration based on deep learning with co-attention matching module.
\newblock {\em Remote Sensing}, 15:3879, 08 2023.

\bibitem{gou2024interpretable}
Shuiping Gou, Xin Wang, Xinlin Wang, and Yunzhi Chen.
\newblock Interpretable matching of optical-sar image via dynamically conditioned diffusion models.
\newblock In {\em Proceedings of the 32nd ACM International Conference on Multimedia}, pages 4358--4367, 2024.

\bibitem{katharopoulos2020transformersrnnsfastautoregressive}
Angelos Katharopoulos, Apoorv Vyas, Nikolaos Pappas, and François Fleuret.
\newblock Transformers are rnns: Fast autoregressive transformers with linear attention, 2020.

\bibitem{FNCC}
J.P. Lewis.
\newblock Fast normalized cross-correlation.
\newblock {\em Ind. Light Magic}, 10, 10 2001.

\bibitem{Reddy1996AnFT}
B.~Srinivasa Reddy and Biswanath~N. Chatterji.
\newblock An fft-based technique for translation, rotation, and scale-invariant image registration.
\newblock {\em IEEE transactions on image processing : a publication of the IEEE Signal Processing Society}, 5 8:1266--71, 1996.

\bibitem{buniatyan2017deeplearningimprovestemplate}
Davit Buniatyan, Thomas Macrina, Dodam Ih, Jonathan Zung, and H.~Sebastian Seung.
\newblock Deep learning improves template matching by normalized cross correlation, 2017.

\bibitem{lin2017featurepyramidnetworksobject}
Tsung-Yi Lin, Piotr Dollár, Ross Girshick, Kaiming He, Bharath Hariharan, and Serge Belongie.
\newblock Feature pyramid networks for object detection, 2017.

\bibitem{huang2021qxssaroptdatasetdeeplearning}
Meiyu Huang, Yao Xu, Lixin Qian, Weili Shi, Yaqin Zhang, Wei Bao, Nan Wang, Xuejiao Liu, and Xueshuang Xiang.
\newblock The qxs-saropt dataset for deep learning in sar-optical data fusion, 2021.

\bibitem{9609993}
Han Zhang, Lin Lei, Weiping Ni, Tao Tang, Junzheng Wu, Deliang Xiang, and Gangyao Kuang.
\newblock Explore better network framework for high-resolution optical and sar image matching.
\newblock {\em IEEE Transactions on Geoscience and Remote Sensing}, 60:1--18, 2022.

\end{thebibliography}

\end{document}